# Generating Chinese Classical Poems with RNN Encoder-Decoder

**Xiaoyuan Yi   and   Ruoyu Li   and   Maosong Sun**
Department of Computer Science & Technology
Tsinghua University
Beijing, China
yxyz2012yxy@163.com liruoyusince1995@gmail.com sms@mail.tsinghua.edu.cn

## Abstract

We take the generation of Chinese classical poem lines as a sequence-to-sequence learning problem, and build a novel system based on the RNN Encoder-Decoder structure to generate quatrains (*Jueju* in Chinese), with a topic word as input. Our system can jointly learn semantic meaning within a single line, semantic relevance among lines in a poem, and the use of structural, rhythmical and tonal patterns, without utilizing any constraint templates. Experimental results show that our system outperforms other competitive systems. We also find that the attention mechanism can capture the word associations in Chinese classical poetry and inverting target lines in training can improve performance.

一声秋雁连天远，(*P*ZPPZ)
The twitter of a wild goose comes from the distant horizon.
万里归帆隔水遥。(*ZPPZZP)
The homebound ships are still ten thousand miles away from the destination
惆怅旧游零落处，(*Z*PPZZ)
I am so sad to be the place where I said goodbye to my travelling companions.
白头萧瑟满江桥。(*P*ZZPP)
There is nothing here, but a gloomy spectacle and the old me in the bridge.

**Figure 1:** A 7-char quatrain generated by our system with the keyword "秋雁" (autumn wild goose) as input. The tone of each character is shown in parentheses. P and Z represent Ping and Ze tones respectively. * means the tone can be either. Rhyming characters are underlined.

## 1 Introduction

Chinese classical poetry is undoubtedly the largest and most bright pearl, if Chinese classical literature is compared to a crown. As a kind of literary form starting from the Pre-Qin Period, classical poetry stretches more than two thousand years, having a far-reaching influence on the development of Chinese history. Poets write poems to record important events, express feelings and make comments. There are different kinds of Chinese classical poetry, in which the quatrain with huge quantity and high quality must be considered as a quite important one. In the most famous anthology of classical poetry, *Three Hundred of Tang Poems* (Sun, 1764), quatrains cover more than 25%, whose amount is the largest.

The quatrain is a kind of classical poetry with rules and forms which mean that besides the necessary requirements on grammars and semantics for general poetry, quatrains must obey the rules of structure and tone. Figure 1 shows a quatrain generated by our system. A quatrain contains four lines, each line consists of seven or five characters. In Archaic Chinese, characters are divided into two kinds according to the tone, namely Ping(level tone) and Ze(oblique tone). Characters of particular tone must be in particular positions, which makes the poetry cadenced and full of rhythmic beauty. Meanwhile, according to the vowels, characters are divided into different rhyme categories. The last character of the first(optional), second and last line in a quatrain must belong to the same rhyme category, which enhances the coherence of poetry.

In this paper, we mainly focus on the automatic generation of quatrains. Actually, the automatic generation of poetry has been a hot issue for a long time, and Deep Learning opens a new door to it now, which makes computer no longer rely on prepared templates, and try to learn the composition method automatically from a large number of excellent poems. Poetry composition by machine is not only a beautiful wish. Based on the poem generation system, interesting applications can be developed, which can be used for education of Chinese classical poetry and the literary researches.

Different from the semantically similar pairs in machine translation tasks, the pair of two adjacent sentences in a quatrain is semantically relevant. We take poem generation as a sequence-to-sequence learning problem, and use RNN Encoder-Decoder to learn the semantic meaning within a single line, the semantic relevance among lines and the use of rhythm jointly. Furthermore, we use attention mechanism to capture character associations to improve the relevance between input lines and output lines. Consisting of three independent line generation blocks (word-to-line, line-to-line and context-to-line), our system can generate a quatrain with a user keyword.

Our system is evaluated based on the task of quatrain generation with two kinds of methods, the automatic and the manual. Experimental results show that our system outperforms other generation systems.

The rest of this paper is organized as follows. In section 2 we will introduce the related methods and systems. In section 3 we will give the details of our system and analyze the advantages of the model. Then Section 4 will give the evaluation experiments design and results. In section 5 we draw a conclusion and point out future work.

## 2 Related Work

The research about poetry generation started in 1960s, and is a focus in recent decades. Manurung(2003) proposed three criteria for automatically generated poetry: grammaticality(the generated lines must obey grammar rules and be readable), meaningfulness(the lines should express something related to the theme) and poeticness(generated poems must have poetic features, such as the rhythm, cadence and the special use of words).

The early methods are based on rules and templates. For example, ASPERA (Gervás, 2001) uses the changes of accent as the templates to fill with words. Haiku generation system (Wu et al., 2009) expands the input queries to haiku sentences according to the rules extracted from the corpus. Such methods are mechanical, which match the requirements of grammaticality, but perform poorly on meaningfulness and poeticness.

One important approach is to generate poems with evolutionary algorithms. The process of poetry generation is described as natural selection. Then through genetic heredity and variation, good results are selected by the evaluation functions (Manurung, 2003; Levy, 2001). However, the methods depend on the quality of the evaluation functions which are hard to be designed well. Generally, the sentences perform better on meaningfulness but can hardly satisfy the poeticness.

Another approach is based on the methods for the generation of other kinds of texts. Yan et al.(2013) generate poems with the method of automatic summarization. While SMT is first applied on the task of couplets generation by (Jiang and Zhou, 2008). They treat the generation of the couplets as a kind of machine translation tasks. He et al.(2012) apply the method on quatrain generation, translating the input sentence into the second sentence, the second one into the third one, and so on. Sentences generated with such method is good at the relevance, but cannot obey the rules and forms.

With the cross field of Deep Learning and Natural Language Process becoming focused, neural network has been applied on poetry generation. Zhang and Lapata (2014) compress all the previous information into a vector with RNN to produce the probability distribution of the next character to be generated.

Our work differs from the previous work mainly as follows. Firstly, we use RNN Encoder-Decoder as the basic structure of our system compared with the method of (He et al., 2012). Moreover, in He's system the rhythm is controlled externally and the results perform poorly on tonal patterns. While our system can learn all these things jointly. Secondly, compared with (Zhang and Lapata, 2014), our model is based on bidirectional RNN with gated units instead of the simple RNN. Besides, Zhang (2014) compress all context information into a small vector, losing much of the useful information. They need two more translation models and another rhythm template to control semantic relevance and tones. While the results of our system can obey these constraints naturally. Finally, we use attention mechanism to capture character associations and invert target sentences in training.

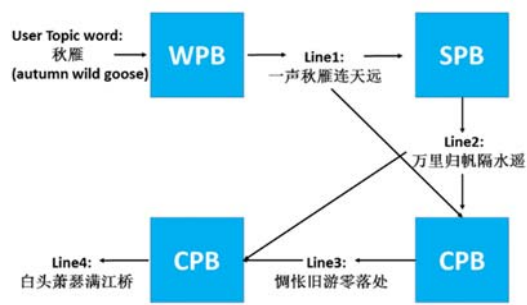

**Figure 2:** An illustration of poem generation process with the topic word "秋雁" as input.

## 3 The Poem Generator

The RNN Encoder-Decoder structure is suitable for sequence-to-sequence learning tasks. In machine translation tasks, the sentence pairs are semantically similar, from which the model learns the corresponding relations. In Chinese classical quatrains, there is a close semantic relevance between two adjacent lines. Such two poem lines are semantically relevant sequence pairs. We use RNN Encoder-Decoder to learn the relevance which is then used to generate a poem line given the previous line. For utilizing context information in different levels, we build three poem line generation blocks (word-to-line, line-to-line and context-to-line) to generate a whole quatrain.

As illustrated in Figure 2, the user inputs a keyword as the topic to show the main content and emotion the poem should convey. Firstly, WPB generates a line relevant to the keyword as the first line. Then SPB takes the first line as input and generates the relevant second line. CPB generates the third line with the first two lines as input. Finally, CPB takes the second and third lines as input and generates the last line.

We train the model and generate poems based on Chinese characters, since there are no effective segmentation tools for Archaic

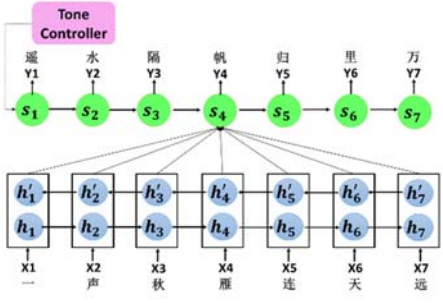

**Figure 3**: An illustration of SPB. The tone controller is used to keep the last character of the second line level-toned

Chinese. Fortunately, the length of Chinese classical poem lines is fixed five or seven characters. And most words in Chinese classical poetry consist of one or two Chinese characters. Therefore, this method is feasible.

### 3.1 Sentence Poem Block (SPB)

SPB (Sentence Poem Block) is used for line-to-line generation. When generating the second line, the only available context information is the first line. Therefore, we use SPB to generate the second line, taking the first line as input. As shown in figure 3, we use bidirectional RNN with attention mechanism proposed by (Bahdanau et al., 2015) to build SPB.

Let us denote an input poem line by $X = (x_1, x_2, \dots, x_{T_x})$, and an output line by $Y = (y_1, y_2, \dots, y_{T_y})$. $e(x_t)$ is the word-embedding of the t-th character $x_t$. $h_t$ and $h'_t$ represent the forward and backward hidden states in Encoder respectively.

In Encoder:

$$d_t = tanh(U[h_{t-1} \cdot r_t] + We(x_t)) \quad (1)$$
$$u_t = \sigma(U_u h_{t-1} + W_u e(x_t)) \quad (2)$$
$$r_t = \sigma(U_r h_{t-1} + W_r e(x_t)) \quad (3)$$
$$h_t = (1 - u_t) \cdot h_{t-1} + u_t \cdot d_t \quad (4)$$
$$g_t = [h_t; h'_t] \quad (5)$$

(1)-(5) are formulas for the computation of forward hidden states. The computation of backward hidden states is similar. • is element-wise multiplication. $g_t$ is the final hidden state of t-th character in Encoder. $r_t$ and $u_t$ are the reset gate and update gate respectively in (Cho et al., 2014). The formulas in decoder are similar. The difference is that a context vector $c_t$ of attention mechanism is used to calculate the hidden state $s_t$ of t-th character in Decoder. $c_t$ is computed as:

$$c_t = \sum_{i=1}^{T_x} \alpha_{t,i} g_t \quad (6)$$

$$\alpha_{t,i} = \frac{\exp(v_{t,i})}{\sum_{j=1}^{T_x} \exp(v_{t,j})} \quad (7)$$

$$v_{t,i} = v_a^T \tanh(W_a s_{t-1} + U_a g_i) \quad (8)$$

Some poems in our corpus are not of high quality. Therefore, we add a neural language model (Mikolov et al., 2010) into the system to improve the meaningfulness and make the poem more idiomatic for modern people. We combine the probability distributions of decoder and language model by:

$$\begin{aligned}P(y_t) &= (1 - \lambda) * f(s_t, y_{t-1}, c_t) \\&+ \lambda lm(y_t | y_{t-1}, \dots, y_1)\end{aligned} \quad (9)$$

Where *lm* is the probability distribution of language model, and $\lambda$ is its weight. Larger $\lambda$ lead to more common lines, and smaller $\lambda$ lead to more novel lines.

Furthermore, as shown in figure 3, when training SPB we invert the target lines for two reasons. First, the final character of second line must be level-tone. While the tail character of third line is oblique-tone. If the final character of input line is level-tone, SPB can't determine the tone of the final character of output line because of the pairs <line2, line3> in training data. We add a tone controller into SPB. Obviously, this control will do harm to meaningfulness of the outputs. Therefore we invert the target sentences in training so that SPB can generate the tail character first, which will minimize the damage on meaningfulness as possible. Furthermore, we find inverting target lines can improve the performance, as the way of inverting the source lines mentioned in (Sutskever et al., 2014).

**3.2 Context Poem Block (CPB)**

To utilize more context information, we build another Encoder-Decoder called CPB (Context Poem Block). The structure of CPB is similar with that of SPB. The difference is that we concatenate two adjacent lines in a quatrain as a long input sequence, and use the third line as the target sequence in training. By this means, the model can utilize information of last two lines when generating current line.

The final characters of the second and the fourth line must rhyme and the final character of the third line must not rhyme. When generating the fourth line, SPB can't determine the rhyme. By taking the second line into consideration, CPB can learn the rhyme. Thus we use CPB to generate the third and the fourth lines. Zhang (2014) utilizes context by compressing all lines into a 200-dimensional vector, which causes a severe loss to semantic information. Whereas our method can save all information. When generating current line, the model can learn focus points with attention, rather than use all context indiscriminately, which will improve semantic relevance between the inputs and the outputs. We don't concatenate more lines for two reasons. Firstly, too long sequences result in low performance. Secondly, relevance between the fourth line and the first line is relatively weak, there is no need to make the system more complicated.

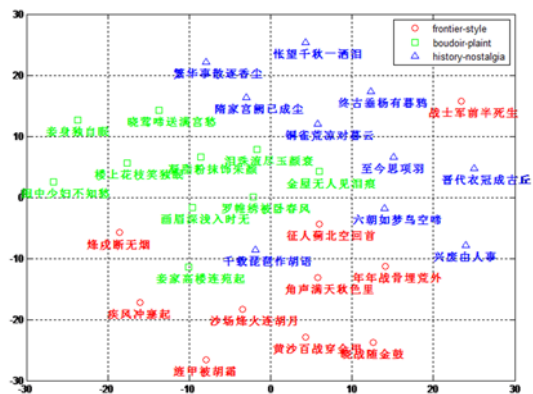

**Figure 4**: 2-D embedding of the learned poem line representations. The circle, square and triangle represent frontier-style, boudoir-plaint and history-nostalgia poem lines respectively.

**3.3 Word Poem Block (WPB)**

A big shortcoming of SMT-based methods is that they need another model to generate the first line. For example, He (2012) expands user keywords, then use constraint templates and a language model to search for a line.

For RNN Encoder-Decoder, words and sentences will be mapped into the same vector space. Since our system is based on characters, words can be considered as short sequences. Ideally, SPB will generate a relevant line taking a word as input. But the training pairs are all long sequences, it won't work well when the input is a short word.

Therefore, we train the third Encoder-Decoder, called Word Poem Block (WPB). Based on the model parameters of trained SPB, we use some <word, line> pairs to train it more to improve WPB's ability of generating long sequences with short sequences.

**3.4 Qualitative Analysis**

Our poetry generation task is on sequence pairs with semantic relevance. We conducted several qualitative experiments. The results show that RNN Encoder-Decoder can capture the semantic meanings and relevance of pairs on such relevance learning tasks well, which is the reason why we use the model to build our system.

| knn | | lcs | |
|---|---|---|---|
| results | cos | results | len |
| 白头萧散满霜风 | 0.45 | 白头萧散满霜风 | 4 |
| 竹风萧瑟满庭秋 | 0.43 | 萧瑟江南庚子山 | 3 |
| 萧瑟西风满玉湾 | 0.40 | 竹风萧瑟满庭秋 | 3 |
| 烟沙萧瑟满汀洲 | 0.38 | 愁红怨白满江滨 | 3 |
| 瑟瑟西风起 | 0.37 | 萧瑟江梅树树空 | 3 |

**Table 1**: The top-5 knn and lcs results of poem line "白头萧瑟满江桥".

### 3.4.1 Poem Line Representations

We used the average of $g_t$ in formula (5) as the representation of a poem line. We selected three types of classical poetry: frontier-style poetry (poetry about the wars), boudoir-plaint poetry (poetry about women's sadness) and history-nostalgia poetry (poetry about history). For each type, we obtained ten lines and used Barnes-Hut-SNE (van der Maaten, 2013) to map their representations into two-dimensional space. As shown in figure 4, lines with the same type gather together. The representations can capture the semantic meanings of poem lines well.

We also generated representations for 20,000 lines then calculated their cos distance and calculated their lcs (longest common sequences) as the comparison.

As shown in table 1, the results of lcs are simple string matchings, but the results of knn have the same meaning as the input's. Though there are no explicit boundaries in input sequences for SPB, the vector representations can contain the information of whole words. For example, "白头"(white hair, a symbol of senility) is a word. In results of knn, the two character "白" and "头" appear as a entirety(or not appear). But for lcs, character "白" appears alone in line "愁红怨白满江滨", in which "白" means white flowers.

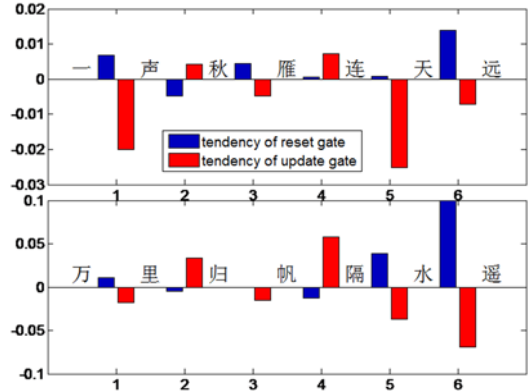

**Figure 5**: An example of word boundaries recognition by gated units. Between every two adjacent characters, the first bar and the second bar show the reset tendency and update tendency. The upper plot is the input line and the lower one is the output line.

### 3.4.2 Gated Units in Word Boundary Recognition

The training and generation are based on characters without any explicit word boundaries in the sequences, but gated units can recognize the word boundaries roughly.

As we can see in formula (2) and formula (3), when $r_t$ tends to be zero and $u_t$ tends to be one, the gated units tend to use current input to update the hidden state. Whereas gated units tend to continue previous hidden states. We used the average value of every elements in $r_t$ as the reset value of t-th character. Along the direction of hidden states propagation, we calculated the difference between reset values of the latter and the previous characters to get the reset tendency. Similarly, we got the update tendency.

As shown in figure 5, higher reset tendency and smaller update tendency mean that the two characters tend to be an entirety. Whereas they tend to be separated. In line "一声秋雁连天远", "一声" and "秋雁" are both words. We can see the reset tendency and update tendency reflect the word boundaries roughly. Furthermore, the tendency of gated units in decoder is similar with that in encoder. This nature makes the vector representations contain information of whole words and makes output lines keep the same structures as input lines.

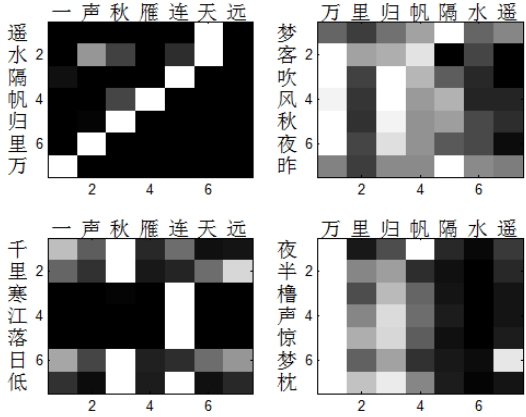

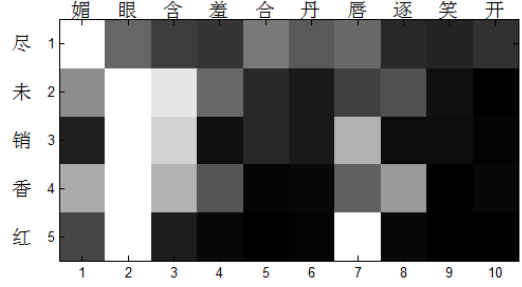

**Figure 7**: An attention example of CPB. The input are two concatenate lines and each line consists of five characters. The output line is inverted.

**Figure 6**: Examples of attention. Each pixel shows the association between two characters. The horizontal lines are inputs and the vertical lines are outputs. Outputs on the top two plots are inverted for the reasons described in section 3.1.

### 3.4.3 Attention Mechanism in Capturing Associations

Different from word alignments in translation task, attention mechanism can capture the implicit associations between two characters. Figure 6 is the visualizations of $\alpha_{t,i}$ in formula (11).

The top two plots show the associations between input and output lines generated by a SPB trained with inverted target lines. The bottom two plots are the results of SPB trained with normal target lines. And in the left two plots, outputs are syntactically similar with inputs. While in the right two ones, output are syntactically different from the inputs.

In the top left plot, when SPB generates the character "帆"(sail), there is a strong dependence on the character "雁"(wild goose). In Chinese classical poetry, "雁" is a symbol of homesickness and "帆" is a symbol of travelers. There is an association between these two characters.

Also, in the top right plot, besides the first character, there are dependencies on the second character "归"(return), since the input line is about home-bound ships and the output line is about the travelers. But the associations are more obvious on pairs with similar syntactic structures.

It is worth mentioning that some researchers (Wang et al., 2015) made an attempt to generate Songci (a kind of Chinese poetry in Song dynasty) with LSTM and drew a conclusion that the attention mechanism is ineffective.

While we got a great improvement by inverting the target lines in training. There are no obvious associations in the bottom two plots compared with results in the top two plots. We think the improvement by inverting may be related to attributive structures in Archaic Chinese. The quantitative evaluation results in section 4 also show that inverting target lines leads to higher scores.

Figure 7 shows an attention result of CPB. As we can see, because the input line is a description of a beautiful woman, attention mechanism focused on two characters, "眼" (eyes) and "唇" (lips). Though there is a color word "丹" (red) in the input line, attention mechanism chose to focus on "眼" and "唇" instead of "丹" for generating the character "红" (also means red color) since in Chinese classical poetry, "红" is often used to describe the beauty of women. Compared with the simple alignments of words with same semantic meanings in translation task, attention mechanism can learn the associations and helps the system to focus on the most relevant information instead of all context, which results in a stronger relevance between input line and output line.

## 4 Experiments

### 4.1 Data and Settings

Our corpus contains 398,391 poems from Tang Dynasty to the contemporary. We didn't use earlier poetry because regular tonal patterns were formed in Tang dynasty. We used 10,000 poems as testing set and other as training set. We extracted three pairs from each quatrain. We used 999,442 pairs to train SPB, and 596,610 pairs to train CPB. For training WPB, we selected 3000 words, and for each words we got 150 lines which the word appears in. Finally we obtained 450,000 word-to-line pairs (half are 5-char and the other half are 7-char). We used the whole training set to train neural language model. We built our system based on GroundHog[1].

### 4.2 Evaluation Design

**BLEU Score Evaluation** Referring to He (2012) and Zhang(2014), we used BLEU-2 score to evaluate the output line given the previous line as input. Since most words in Chinese classical poetry consist of one or two characters, BLEU-2 is effective. It's hard to obtain human-authored references for poem lines so we used the method in (He et al., 2012) to extract references automatically. We selected 4,400 quatrains from testing set (2,200 of them are 5-char and other 2,200 are 7-char) and extracted 20 references for each line in a quatrain(except the last line). We compared our system with the system in (He et al., 2012).

**Human Evaluation** Since poetry is a kind of creative text, human evaluation is necessary. Referring to the three criteria in (Manurung, 2003), we designed five criteria: Fluency (are the lines fluent and well-formed?), Coherence(does the quatrain has consistent topic across four lines?), Meaningfulness(does the poem convey some certain messages?), Poeticness(does the poem have poetic features such as the poetic images?), Entirety(the reader's general impression on the poem). Each criterion was scored from 0 to 5.

We compared four comparison systems. **PG**, our system. **SMT**, He (2012)'s system[2]. **DX**, the DaoXiang Poem Creator[3]. This system is the pioneer for Chinese classical poetry generation. It has been developed for 15 years and been used over one hundred million times. **Human**, the poems of famous ancient poets containing the given keywords.

We selected 24 typical keywords and generated two quatrains (5-char and 7-char) for each keyword using the four systems[4]. By this means, we obtained 192 quatrain (24*4*2) in total. We invited 16 experts[5] on Chinese classical poetry to evaluate these quatrains. Each expert evaluated 24 quatrains.

---

[1] https://github.com/lisa-groundhog/GroundHog.
[2] http://duilian.msra.cn/jueju/.
[3] http://www.poeming.com/web/index.htm.
[4] Because the keywords is limited in SMT, we used lines of ancient poets for the keywords not included in their list as the first lines.
[5] All the experts have the ability to assess and create Chinese classical poetry.

The 16 experts were divided into two groups and each group completed the assessments of the 192 poems. Thus we got two scores for each quatrain and we used the average score.

| Models | Line2 | | Line3 | | Line4 | | Average | |
|---|---|---|---|---|---|---|---|---|
| | 5-char | 7-char | 5-char | 7-char | 5-char | 7-char | 5-char | 7-char |
| **SMT** | 0.526 | 0.406 | 0.262 | 0.214 | 0.432 | 0.314 | 0.407 | 0.311 |
| **SPB0** | 0.865 | 1.163 | 0.910 | 1.229 | 1.046 | 2.167 | 0.940 | 1.520 |
| **SPB0+attention** | 0.773 | 0.956 | 0.478 | 0.728 | 0.831 | 1.450 | 0.694 | 1.045 |
| **SPB0+attention+src invert** | 0.739 | 1.048 | 0.671 | 1.049 | 0.876 | 1.453 | 0.762 | 1.183 |
| **SPB0+attention+trg invert** | **1.126** | **1.900** | **1.251** | **1.441** | **1.387** | **2.306** | **1.255** | **1.882** |

**Table 2**: BLEU-2 scores on quatrains. SPB0 is the simple structure without attention mechanism or inverting target lines.

| Models | Fluency | | Coherence | | Meaningfulness | | Poeticness | | Entirety | |
|---|---|---|---|---|---|---|---|---|---|---|
| | 5-char | 7-char | 5-char | 7-char | 5-char | 7-char | 5-char | 7-char | 5-char | 7-char |
| **SMT** | 1.65 | 1.56 | 1.52 | 1.48 | 1.42 | 1.33 | 1.69 | 1.56 | 1.48 | 1.42 |
| **DX** | 2.53 | 2.33 | 2.19 | 1.96 | 2.31 | 2.00 | 2.52 | 2.31 | 2.29 | 2.08 |
| **PG** | 3.75 | 3.92 | 3.42 | 3.48 | 3.50 | 3.50 | 3.50 | 3.67 | 3.60 | 3.67 |
| **Human** | **3.92** | **3.96** | **3.81** | **4.10** | **4.08** | **4.13** | **3.75** | **4.02** | **3.96** | **4.21** |

**Table 3**: Human evaluation results. The Kappa coefficient of the two groups' scores is 0.62. Since generations of PG and SMT are both sequence-to-sequence, we invited another expert to select a best line from the top-10 list in the generation of each line.

**4.3 Evaluation Results**

Table 2 shows the BLEU-2 scores. Because DX system generates poetry as a whole, we only compared our system with SMT on single line generation task. Given an input line, the output of SMT is often the original adjacent next line of ancient poets. To be fair, we removed the lines of ancient poets from the top-n lists of our system and SMT, then we used the best ones. In Chinese classical poetry, the relevance between two lines in a pair is related to the position. Therefore He et al.(2012) use pairs in different positions to train corresponding position-sensitive models. Because of the limited training data, we used pairs in all positions to train SPB. Even so, we got much higher BLEU scores than SMT in all positions[6]. Moreover, 95% of the outputs generated by our system observe tonal constraints, but only 31% of SMT's outputs observe the constraints.

We can also see that simply adding the attention mechanism to system made against the performance, but the combination of attention and inverting led to a great improvement. In (Sutskever et al., 2014), they find reversing source sentences can improve the LSTM's performance and conclude that this preprocessing can introduce short term dependencies between the source and the target sentences which will make the optimization problem easier. As their explanation, inverting source sentences and inverting target sentences should be equivalent. But as shown in table 2, inverting source sentences made little improvement in our task. We think this is because of the attributive structure in Chinese classical poetry. There are many words with the structure *attributive + central word* in Chinese poetry. For example, "青草"(green grass), the character "青" (green) is attributive and the character "草" (grass) is central word. In normal generation order, "青" will be generated earlier than "草". But there are many other characters can be the central word of "青", such as "青山" (green mountain), "青云"(green cloud), "青烟" (green smoke), which increases the uncertainty in the generation of "草". Whereas inverting target sentence can reduce this uncertainty since the attributive of "草" is often "青" in Chinese classical poetry. When generating t-th character, the only information can be used to determine the attention weights $\alpha_{t,i}$ by Decoder is $s_{t-1}$, thus reducing uncertainty will lead to better attention weights. This is why simply adding the attention mechanism didn't work.

---

[6] We got lower BLEU-1 scores of SMT system than those reported in (He et al., 2012), because we removed the lines of ancient poets from candidate lists.

As shown in table 3, our system got higher scores than other systems, expect the Human. For SMT and DX, the scores of 7-char poems are lower than that of 5-char poems in all criteria (both in Human evaluation and BLEU evaluation) because the composition of 7-char quatrains is more difficult. But the poems generated by PG got higher scores on 7-char poems, benefiting from gated units and attention mechanism. Scores of PG is closed to scores of Human, though there is still a gap.

We also asked the experts to select a best line from the quatrains evaluated. 37% of the selected 5-char lines are of our system and 42% are of poets. And 45% of the selected 7-char lines are of our system, 45% are of poets. This indicates that our system has little difference with poets, at least in generating meaningful sentences.

| Blocks | 5-char | | 7-char | |
|---|---|---|---|---|
| | Line3 | Line4 | Line3 | Line4 |
| SPB | 0.603 | 0.576 | 0.536 | 0.478 |
| **CPB** | **0.614** | **0.732** | **0.573** | **0.713** |
| A3 | - | 0.696 | - | 0.621 |

**Table 4**: Generation probability ranking scores.

Though the qualitative experiments have shown the feasibility of our system, we also conducted a quantitative experiment to compare SPB and CPB. We trained another Encoder-Decoder called A3, which concatenates three adjacent lines as input and takes the fourth line as target. BLEU is used to evaluate the line-to-line generation task which focuses on the semantic relevance between two lines, but CPB and A3 both concentrate on the relevance between the generated line and all context. Thus BLEU is not suitable for comparing these three Blocks.

Therefore, we designed another comparison method, called generation probability ranking score (GPRS). Taking a sequence as input, we can get the generation probability of a specified line by Decoder. We used the testing data in section 4.2. By inputting a testing line to a Block, we got the generation probabilities of all references (including the original next line of ancient poet). We sorted all the references descending on generation probability. Then we calculated the GPRS of the input line by $\frac{N-1-r}{N-1}$, r is the rank of the original next line of ancient poet ($0 \leq r < N$) and N is the number of references. Higher GPRS indicates that the Block has a greater chance to generate the original lines, which means the Block can capture the relevance between generated lines and context better.

As shown in table 4, we can see that CPB got the highest scores. And A3 is better than SPB but worse than CPB, for the reasons we mentioned in section 3.2.

梦断中秋月，
When I woke up from the dream suddenly, I saw the mid-autumn moon.
天寒咽暮蝉。
It was too cold for the cicadas to sing in the evening.
不堪送归客，
I couldn't bear the pain of seeing my friends off.
寂寞对床眠。
The only thing I could do was trying to fall asleep with loneliness.

谁怜两地中秋月，
Who will feel sympathy for the separated us? Only the autumn moon will.
独照西窗一夜凉。
The moonlight through the window is so lonely on the cold night.
行到故园应怅望，
Maybe your are overlooking and trying to find where I am in the distance.
哀词遗恨满潇湘。
While I can only put my missing and sadness in my poems, and let the melancholy fill the Xiao River and the Xiang River.

**Figure 8:** Two poems generated by our system with keyword "秋月" (autumn moon) as input.

## 5 Conclusion and Future Work

In this paper, we take the generation of poem lines as a sequence-to-sequence learning problem, and build a novel system to generate quatrains based on RNN Encoder-Decoder. Compared with other methods, our system can jointly learn semantic meanings, semantic relevance, and the use of rhythmical and tonal patterns, without utilizing any constraint templates. Both automatic evaluation and human evaluation show that our system outperforms other systems, but there is still a gap between our system and ancient poets.

We show that RNN Encoder-Decoder is also suitable for the learning tasks on semantically relevant sequences. The attention mechanism can captures character associations, and gated units can recognize word boundaries roughly. Moreover, inverting target lines in training will lead to better performance.

There are lots to do for our system in the future. Based on different blocks, our system is extensible. We will improve our system to utilize more context information and generate other types of Chinese poetry, such as Songci and Yuefu. We also hope our work could be helpful to other related work, such as the building of poetry retrieval system, word segmentation of poems, and literature researches.